\documentclass[10pt,twocolumn,letterpaper]{article}

\usepackage[pagenumbers]{wacv} 

\usepackage[accsupp]{axessibility}
\usepackage{amsfonts}
\usepackage{makecell}
\usepackage{stackengine}
\usepackage{multirow}
\usepackage[dvipsnames]{xcolor}
\usepackage[boldmath]{numprint}

\usepackage[utf8]{inputenc}
\usepackage{amsmath}
\usepackage[ruled,vlined]{algorithm2e}

\usepackage{algorithmic}
\usepackage{placeins}
\usepackage{balance}
\usepackage{float}

\definecolor{wacvblue}{rgb}{0.21,0.49,0.74}
\usepackage[pagebackref,breaklinks,colorlinks,allcolors=wacvblue]{hyperref}

\def\wacvPaperID{} 
\def\confName{WACV}
\def\confYear{2026}

\title{Chain-of-Look Spatial Reasoning for Dense Surgical Instrument Counting}

\author{Rishikesh Bhyri \and Brian R Quaranto \and Philip J Seger \and Kaity Tung \and Brendan Fox \and Gene Yang \and Steven D. Schwaitzberg \and Junsong Yuan \and Nan Xi \textsuperscript{\textdagger} \and Peter C W Kim \textsuperscript{\textdagger}  \and \mbox{} \\
State University of New York at Buffalo\\
{\tt\small \{rbhyri, brianqua, pseger, kaitytun, btfox, geneyang, schwaitz, jsyuan, pckim, nanxi\}@buffalo.edu} \\
\textsuperscript{\textdagger} {\tt\small Equal Advising}
}

\begin{document}

\twocolumn[{
\renewcommand\twocolumn[1][]{#1}%
\maketitle
\begin{center}
    \centering
    \captionsetup{type=figure}
    \includegraphics[width=0.97\textwidth]{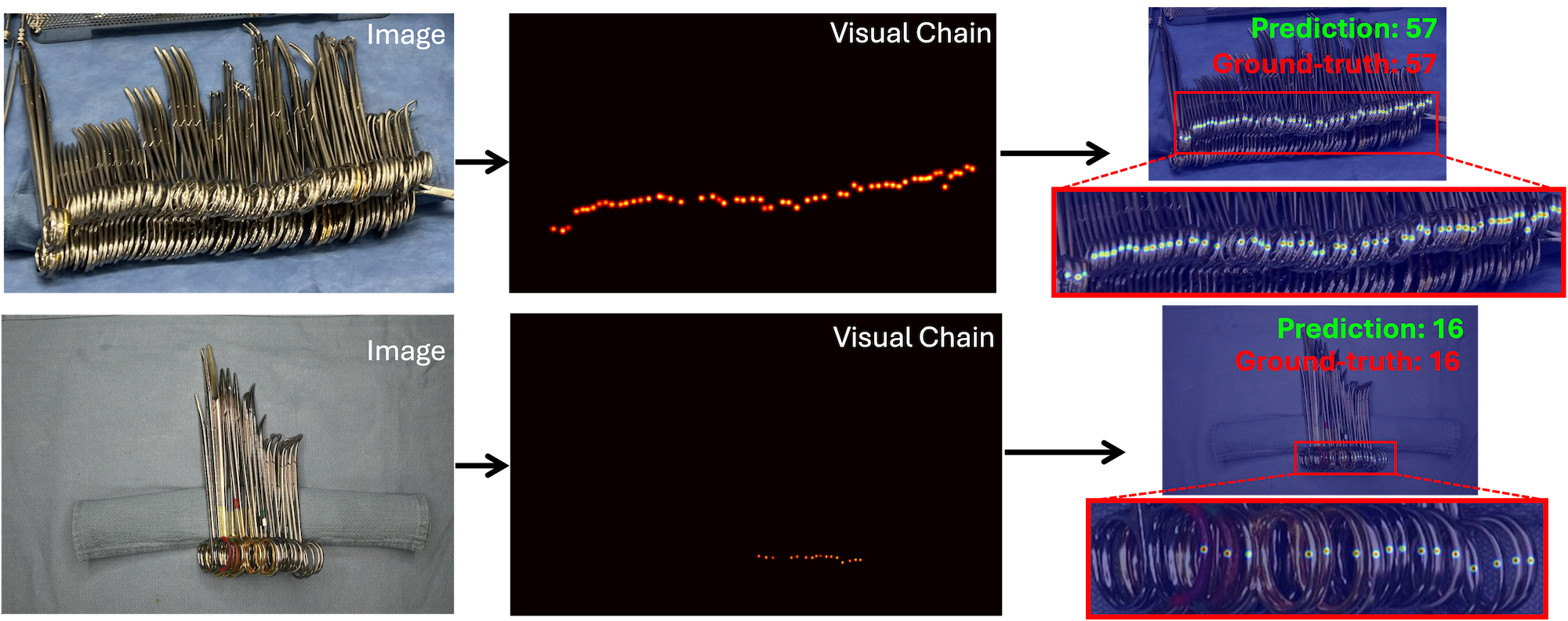}
    \vspace{-2mm}
    \captionof{figure}{\textbf{High-density surgical instrument counting.} Counting surgical instruments reliably in high density scenarios is challenging due to severe visual clutter and tight spatial packing of objects. To improve robustness, we propose Chain-of-Look Spatial Reasoning to introduce \textit{visual chains} into the counting process, explicitly modeling the sequential characteristic of human visual counting. In the above figure, the first column indicates original high-density surgical instrument images, the second column presents visual chains and the third column shows the predicted counting results, where detected surgical instrument handles are highlighted with laser points.}
    \label{fig:Fig1}
\end{center}%
}]

\begin{abstract}
Accurate counting of surgical instruments in Operating Rooms (OR) is a critical prerequisite for ensuring patient safety during surgery. Despite recent progress of large visual-language models and agentic AI, accurately counting such instruments remains highly challenging, particularly in dense scenarios where instruments are tightly clustered. To address this problem, we introduce Chain-of-Look, a novel visual reasoning framework that mimics the sequential human counting process by enforcing a structured visual chain, rather than relying on classic object detection which is unordered. This visual chain guides the model to count along a coherent spatial trajectory, improving accuracy in complex scenes. To further enforce the physical plausibility of the visual chain, we introduce the neighboring loss function, which explicitly models the spatial constraints inherent to densely packed surgical instruments. We also present SurgCount-HD, a new dataset comprising 1,464 high-density surgical instrument images. Extensive experiments demonstrate that our method outperforms state-of-the-art approaches for counting (e.g., CountGD, REC) as well as Multimodality Large Language Models (e.g., Qwen, ChatGPT) in the challenging task of dense surgical instrument counting. The code and dataset is available at {\small \url{https://github.com/rishi1134/CoLSR.git}}
\end{abstract}
    
\vspace{-2mm}
\section{Introduction}
\label{sec:intro}
\vspace{-2mm}
Counting surgical objects including instruments before and after a surgical procedure is a critical safety protocol in Operating Rooms (OR), aimed at preventing retained surgical items. Despite its importance, this process is still predominantly performed manually by surgical staff, making it time-consuming, labor-intensive, and prone to human error, particularly in high-density settings where instruments are closely clustered or visually occluded. These challenges are further exacerbated under time pressure or in emergency procedures. Given that the average operating room costs approximately \$100 per minute, delays caused by manual counting can have significant financial implications in addition to clinical risks. Thus, automating the surgical instrument counting process holds great potential to reduce the workload on surgical teams, minimize human errors, and enhance workflow efficiency and patient safety. However, accurate automated counting remains a challenging task due to visual complexity and high similarity among instruments in real-world OR environments.

Most existing approaches to object counting fall into two main categories: density map-based and detection-based methods. Density map-based methods estimate object counts by summing the predicted density values across an image, while detection-based methods count the number of predicted bounding boxes. Although these methods have achieved strong performance in various scenarios such as crowd counting and open-set counting, they fundamentally treat counting as a set-based problem, ignoring the sequential nature of how humans count, particularly in complex, high-density environments. In practice, humans typically follow a consistent visual path when counting objects to avoid omissions or duplications, and to verify the counting. For instance, as shown in Figure~\ref{fig:Fig1}, technicians and nurses counting the surgical instruments tend to follow a structured visual sequence, such as scanning from left to right or right to left. This sequential reasoning process is crucial in ensuring accurate counts under cluttered and visually challenging conditions, yet it is largely overlooked in existing automated counting frameworks.

Motivated by the importance of sequential visual reasoning in human counting behavior, we introduce the \textbf{Chain-of-Look Spatial Reasoning} (CoLSR) framework that explicitly models the counting sequence in dense object scenes by locating each object as a counting point. CoLSR \textit{explicitly models the sequential nature of human counting}, which is particularly critical in \textit{complex, high-density environments}. Introducing a direction for counting is especially important in high-stakes surgical scenarios, where medical staffs always follow a \textbf{strict counting direction} to ensure accuracy rather than counting instruments in a random order. Unlike traditional methods that treat object instances independently, our CoLSR not only predicts the locations of target objects (e.g., the handles of surgical instruments in our task) but also captures the \textit{spatial dependencies} and \textit{structural relationships} among them. 
To achieve this, we first generate visual chains using a transformer-based counting model, CountGD ~\cite{amini2024countgd}. These visual chains provide guidance for our model, enabling it to reason spatially and improve prediction accuracy. To further align the predicted visual chains with the underlying spatial structure of the scene, we introduce a novel neighboring loss that encourages the predicted object order to match the ground-truth sequence. The neighboring loss encourages the model to consider the proximity and ordering of adjacent objects, and enforces consistency with realistic spatial arrangements by penalizing implausible gaps or overlaps. Therefore, the neighboring loss guides the model to learn a coherent spatial chain that mirrors the sequential patterns humans naturally follow during counting, leading to more robust and accurate performance in high-density scenarios. 

We evaluate our CoLSR framework through extensive experiments on a high-density surgical instrument dataset that we construct. Empirical results demonstrate that CoLSR consistently outperforms state-of-the-art (SOTA) object counting methods and multimodality large language models in the context of densely packed surgical instruments, highlighting its effectiveness in real-world, high-complexity scenarios.

In summary, our contributions include:
\begin{itemize}
    \item We introduce the novel and challenging task of dense surgical instrument counting, a problem with significant clinical implications. 
    \item We introduce the Chain-of-Look Spatial Reasoning framework that incorporates \textit{visual chains} into the counting process, explicitly modeling the sequential nature of human visual counting. We design the neighboring loss to equip the model with spatial reasoning capabilities by enforcing inter-object relationships and realistic spatial constraints.
    \item We construct a comprehensive dataset comprising 1,464 high-density surgical instrument images collected from diverse real-world clinical settings. Extensive experiments show that CoLSR delivers significant improvements over existing methods for high-density surgical instrument counting, achieving both high accuracy and fast inference.
\end{itemize}
\section{Related Work}

\begin{figure*}[t]
\centering
\includegraphics[width=0.98\textwidth]{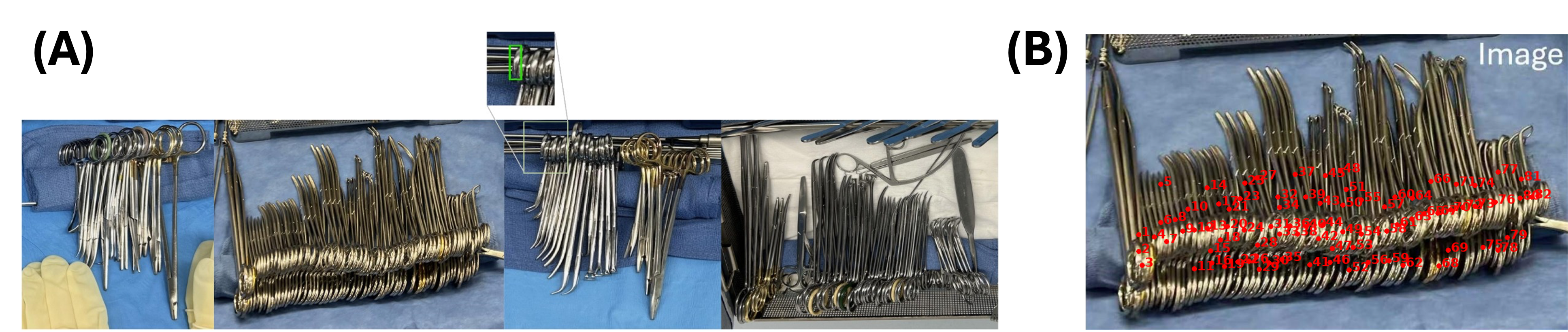} 
\vspace{-1mm}
\caption{(A) \textbf{Representative images from the SurgCount-HD dataset.} Sample images from the dataset, showing typical variations and an example annotation. (B) \textbf{Test result from GPT5.} We evaluate GPT-5 on an example from our SurgCount-HD dataset, where detected surgical instruments are highlighted with \textcolor{red}{red} dots. GPT-5 predicts a count of 84, whereas the ground truth is 57.}
\label{fig:representative_images_GPT5}
\end{figure*}
\vspace{-1mm}

\subsection{Object Counting}
\vspace{-1mm}
\noindent \textbf{Human Counting.} The spatial strategies and typical scanning patterns humans employ when counting stationary objects are deeply rooted in the psychology of enumeration. Gelman et al. \cite{gelman2009child} highlight the principle of one-to-one correspondence (each item gets one and only one tag) which strongly implies a systematic way of going through a set to ensure accuracy. Logan et al. \cite{logan2008eyes} further demonstrate that the number of eye movement fixations increases linearly with the number of objects during counting, indicating a sequential, item-by-item processing strategy when dealing with larger quantities. 

\noindent \textbf{Machine Counting.} Object counting with machines has traditionally been a detection-based or density-based estimation problem. The density-based approach is particularly prevalent in crowd-counting applications \cite{guo2024regressor,peng2024singledomaingeneralizationcrowd,siu1999neural,kong2006viewpoint,Lempitsky2010learning,marana1997estimation,Xie04052018} mainly due to its dense and cluttered scenes. Prior work in density-based counting has shown superior counting accuracy \cite{arteta2014interactive,arteta2016counting,amininaieni2023openworldtextspecifiedobjectcounting,djukic2023lowshotobjectcountingnetwork,liu2023countrtransformerbasedgeneralisedvisual} compared to detection-based methods, which rely on bounding box prediction, a process that becomes challenging in the presence of overlapping and occluded object boundaries. Recent detection-based approaches \cite{amini2024countgd,Dai_2024_CVPR,lin2018focallossdenseobject} have begun to achieve improved accuracy with the advantage of the inherent object localization capability. Recent works like DAVE \cite{pelhan2024davedetectandverifyparadigm}, DQ-DETR \cite{huang2024dqdetrdetrdynamicquery}, and CAD-GD \cite{Wang_2025_CVPR} leverage both detection- and density-based approaches to achieve better counting and localization accuracy.

Although CountGD demonstrates strong performance in open-set object counting, its category-agnostic design limits semantic specificity and dense object localization. However, its open-world setup enables generation of ‘good-enough’ priors even in few-shot settings, making it a useful guiding mechanism for class-specific models with reduced data and computational overhead.

\vspace{-1mm}
\subsection{Chain-of-Look Prompting}
\vspace{-2mm}
Chain-of-thought prompting \cite{xi2023chain, xi2023open, wei2022chain} has been shown to enhance the reasoning capabilities of Large Language Models (LLMs) and reduce hallucinations by breaking down the task into smaller steps in LLMs. Drawing a parallel to vision domain, we adopt the CoL prompting strategy to support spatial reasoning and model the sequential nature of counting by prompting the detected visual cues in a chained fashion, thereby enabling more structured visual attention across densely packed instruments.
\vspace{-1mm}

\subsection{Prompt Tuning}
\vspace{-2mm}

Parameter-Efficient Fine-Tuning (PEFT) approaches, such as prompt tuning \cite{jia2022visualprompttuning}, have proven effective in adapting to newer data distributions while reducing both data and computational requirements. Beyond efficiency, Yao et al. \cite{yao2024sepselfenhancedprompttuning} demonstrate that fusing frozen tokens with learnable prompts further boosts the generalization and discriminative capability of prompt tuning. Moreover, Kang et al. \cite{kang2024vlcounter} introduce semantic-conditioned prompts that guide the image encoder toward extracting target-semantic-highlighted visual features.

With these existing contributions in mind, we explore integrating such PEFT approaches with the existing CountGD framework to address its limitations in handling out-of-distribution classes, particularly in highly dense scenarios. 

\section{Surgical Instrument Counting Dataset}
\vspace{-2mm}

We introduce SurgCount-HD, a novel dataset consisting of \textbf{H}igh-\textbf{D}ensity arrangements of \textbf{Surg}ical instruments collected prior to surgical procedures. Each image contains various types of surgical instruments compactly organized on the back table (a common surgical preparation surface). The dataset focuses on instrument layouts where handles are oriented toward the camera, and bounding-box annotations are provided for these handles, as shown in Figure~\ref{fig:representative_images_GPT5} (A). Translational and rotational augmentations were applied and the final the dataset comprises 1,236 training images and 228 test images. All images were resized such that the shorter edge is scaled to 800 pixels while preserving the original aspect ratio. All annotations represent a single class, namely ``\textit{circular instrument handle}''.

We used Roboflow \cite{roboflowPlatform} platform to manually label instrument handles across densely packed scenes. The annotation process required substantial manual effort due to the high density and visual similarity among instruments. The data collection and annotation process spanned several months and involved multiple domain experts to ensure accuracy and consistency. 

The SurgCount-HD dataset presents significant challenges due to the tightly clustered and visually occluded surgical instruments. Even for human annotators, counting in such high-density scenarios is time-consuming and labor-intensive. To assess the difficulty of this dataset, we evaluated GPT5 \cite{gpt5} on selected examples from SurgCount-HD. As shown in Figure~\ref{fig:representative_images_GPT5} (B), GPT5 performs poorly in these dense settings (detected 84 instruments, where the ground-truth is 57), highlighting the challenge of this SurgCount-HD dataset.

\section{Chain-of-Look Spatial Reasoning}
\label{sec:arch}
\begin{figure*}[t]
\centering
\includegraphics[width=0.7\textwidth]{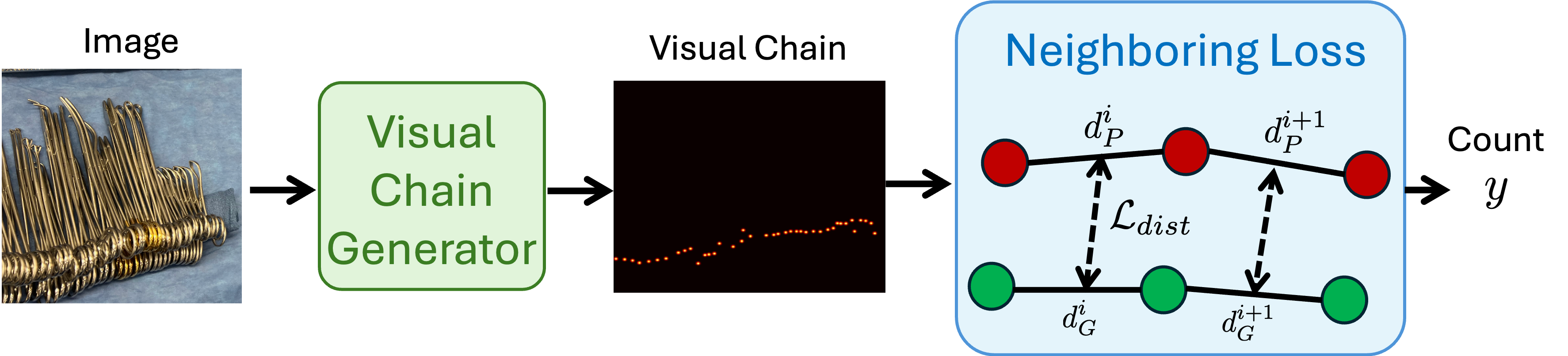} 
\vspace{-2mm}
\caption{\textbf{Architecture of Chain-of-Look Spatial Reasoning framework.} High density surgical instrument images are first fed into visual chain generator to produce visual chains. Neighboring loss is further applied to guide the counting process following the visual chain.}
\label{fig:overall_arch}
\end{figure*}
\subsection{Problem Formulation}
In this section, we introduce the Chain-of-Look Spatial Reasoning (CoLSR) framework for high density surgical instrument counting. Given an image $\mathbf{I} \in \mathbb{R}^{H \times W \times 3}$ of high density surgical instruments, our goal is to train a model $\mathcal{F}_{\theta}$ to localize surgical instrument handles and count the number $y$ of surgical instruments in the image based on the localized instrument handles: $y = \mathcal{F}_{\theta} (I)$, where $\theta$ denotes the parameters of our model, $H$ and $W$ represents the height and weight of the given image. Two major components construct the CoLSR framework: (1) \textit{Visual Chain Generator} for producing the visual chain to guide the counting process; (2) \textit{Neighboring Loss Function} to introduce physical constraints along the visual chain. 
\vspace{-2mm}
\subsection{Visual Chain Generator}
As shown in Figure~\ref{fig:overall_arch}, the first part of CoLSR is to generate visual chain of the given image. The visual chain serves as a \textit{structured visual sequence} to guide the model's counting process under cluttered and visually challenging conditions. 

\begin{figure*}[t]
\centering
\includegraphics[width=0.87\textwidth]{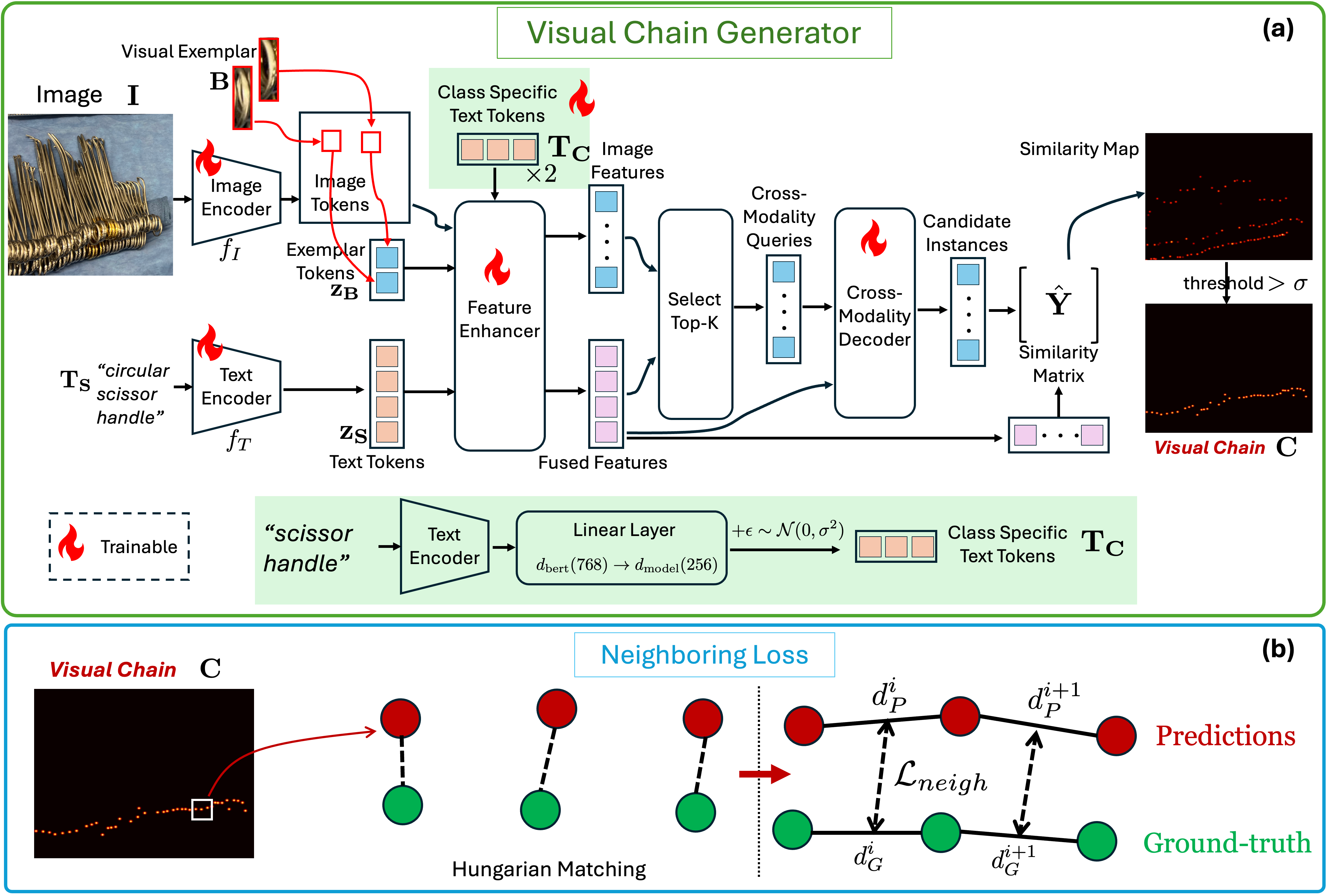} 
\vspace{-2mm}
\caption{\textbf{Visual Chain Generator and Neighboring loss function.} (a) Detailed architecture of Visual Chain Generator; (b) Neighboring loss and Distance loss. Detailed illustrations on the architecture can be found in Section~\ref{sec:arch}.}
\label{fig:detail_arch}
\end{figure*}

The visual chain generator is constructed based on the CountGD model \cite{amini2024countgd}. Different from CountGD, we also take class-specific text tokens as input to enhance the quality of generated visual chain. Figure~\ref{fig:detail_arch} (a) depicts the detailed architecture of Visual Chain Generator. 

\noindent \textbf{Image Encoder.} We first encode the input image $\mathbf{I}$ with a Swin-B version of Swin Transformer \cite{liu2021swin} based Image Encoder $f_I$ into spatial feature maps at three different scales, followed by $1\times 1$ convolution to produce image tokens $\mathbf{z_I}$ of $256$ dimensions. The visual exemplar tokens $\mathbf{z_B}$ are obtained from the image tokens using aligned region-of-interest pooling (RoIAlign) with the pixel coordinates specified by the visual exemplars $\mathbf{B}$. The generated visual exemplar tokens also have $256$ dimension, which is the same with image tokens and text tokens. 

\noindent \textbf{Text Encoder.} For text input, a BERT-based text transformer \cite{devlin2019bert} encoder $f_T$ is employed to encode the text description $\mathbf{T_S}$ into a sequence of tokens $\mathbf{z_T}$ with at most $256$ dimensions. Then the $n$ image tokens, $p$ visual exemplar tokens and $q$ text tokens are applied with the feature enhancer $f_{\phi}$. 

\noindent \textbf{Feature Enhancer.} The generated visual exemplar tokens $\mathbf{z_B}$ is fused with the text tokens $\mathbf{z_T}$ through the feature enhancer $f_{\phi}$ with 6 blocks self-attention modules. The generated fused feature $\mathbf{z_{B,T}}$ are further fused with the image tokens $\mathbf{z_I}$ through the feature enhancer $f_{\phi}$ with 6 blocks cross-attention modules. To enhance the grounding ability of our model, we take the prompt tuning approach to introduce class-specific text tokens $\mathbf{T_C}$ as additional inputs for the feature enhancer (see supplementary for details). These class-specific text tokens serve as learnable parameters to further improve the results of generated visual chains. Therefore, the outputs from the feature enhancer are computed as
\begin{equation}
    \begin{split}
         \mathbf{z_{B,T}, \mathbf{z_I}} = &f_{\phi} (( f_{\theta}(\mathbf{X}), \\ 
         &\text{RoIAlign}(f_{\theta}(\mathbf{X}), \mathbf{B}), f_{T}(\mathbf{T_S}), f_T (\mathbf{T_C}) ).
    \end{split}
\end{equation}

\noindent \textbf{Query Selection.} $k$ image patch tokens are selected which achieve the highest similarity with the fused visual exemplar $\mathbf{B}$ and text description $\mathbf{T_S}$. Following CountGD, we set $k$ to $900$, serving as cross-modality queries input to the cross-modality decoder $f_{\psi}$. 

\noindent \textbf{Cross-modality Decoder.} The cross-modality decoder $f_{\psi}$ contains 6 blocks of self-attention and cross-attention to enhance the cross-modality queries. The final output of confidence score $\hat{\mathbf{Y}}$ is computed as 
\begin{equation}
\label{eq:y_hat}
    \hat{\mathbf{Y}} = \text{Sigmoid} ( f_{\psi} ( \mathbf{z_I}, \mathbf{z_{B,T}}, f_{S}(\mathbf{z_I}, \mathbf{z_{B,T}}^T, k) ) \mathbf{z_{B,T}}^T ),
\end{equation} 
where $f_{S}$ denotes the above Query Selection module. 

\subsection{Neighboring Loss}
The generated visual chain provides a coarse estimate of the instrument count and serves as a structural guidance to guide our model toward precise surgical instrument counting. To incorporate directional consistency along the visual chain, we introduce a neighboring loss term into the training objective. As illustrated in Figure~\ref{fig:detail_arch} (b), given the predicted visual chain $\textbf{C}$, we first use the Hungarian matching algorithm to associate each predicted bounding box with its corresponding ground-truth bounding box. Specifically, for the predicted $\{b_i \}_{i=1}^{N_P}$ and ground-truth surgical instrument handle $\{b_j \}_{j=1}^{N_G}$, the value function $\mathbf{v}$ for Hungarian matching algorithm is defined as:
\begin{equation}
    \mathbf{v}_{i,j} = d_{i,j} + \ \ \mathcal{L}_{i,j}^{cls},
\end{equation}
where $\mathbf{v}_{i,j}$ is the value function for the pair $(i, j)$ from predictions and ground-truth, $d_{i,j}$ indicates the $l_1$ norm of the center points and $\mathcal{L}_{cls}$ denotes the classification cost (see supplementary for further details). 

Given the matched bounding boxes, we examine the local regions of detected surgical instrument handles in a fixed direction (either left-to-right or right-to-left). As illustrated in Figure~\ref{fig:detail_arch} (b), we introduce a neighboring loss that encourages the distances between adjacent center points of bounding boxes in the predictions to closely match those in the ground truth: 
\vspace{-1mm}
\begin{equation}
    \mathcal{L}_{neigh} = \sum_{i=1}^N || d_P^i-d_G^i ||_2, 
\end{equation}
where $d_P^i$ denotes the distance between two neighboring center points of predicted bounding boxes, $d_G^i$ indicates the distance between two counterpart neighboring center points of ground-truth bounding boxes. This neighboring loss function promotes spatial consistency in the ordering of instruments and enforces a visual chain structure in the model's reasoning process, enabling the Chain-of-Look mechanism. We further discuss this effect in Section 1 of the Supplement Materials.

\subsection{Training}
\vspace{-2mm}
As shown in Figure~\ref{fig:detail_arch} (a), we train the image encoder $f_I$, text encoder $f_T$, feature enhancer $f_{\phi}$, cross-modality decoder $f_{\psi}$ and the learnable class specific text tokens $\mathbf{T_C}$. The optimization objective of the whole model includes CountGD \cite{amini2024countgd}'s original bounding box localization loss, classification loss, and our proposed neighboring loss:
\vspace{-1mm}
\begin{equation}
\label{eq:lossfunc}
\begin{split}
    \mathcal{L} &= \lambda_{loc} \mathcal{L}_{loc} + \lambda_{neigh} \mathcal{L}_{neigh} + \lambda_{cls}\mathcal{L}_{cls}\\
    &=\lambda_{loc} \sum_{i=1}^{N_G} |\hat{c}_i - c_i| + \lambda_{neigh} \sum_{i=1}^{N_G} || d_P^i-d_G^i ||_2 \\ &+\lambda_{cls}FocalLoss(\hat{\mathbf{Y}}, T),
\end{split}
\end{equation}
where $\lambda_{loc}$, $\lambda_{cls}$ and $\lambda_{neigh}$ are weights to control each loss term, $\hat{\mathbf{Y}}$ is the similarity matrix from Equation \ref{eq:y_hat} and $T \in \{0, 1\}^{N_P \times (N_G+1)}$ denotes the optimal Hungarian matching between the $N_P$ predicted queries and the $N_G$ ground truth handle instances, including an additional label for ``no object'' similar to CountGD.

During training, the model receives a high density surgical instrument image \( \mathbf{I} \) along with visual exemplars \( \mathbf{B} \) as inputs. These inputs are processed through the image encoder and cross-modality modules to generate query representations, which are then optimized using the loss functions.

\vspace{-2mm}
\subsection{Inference}
\vspace{-1mm}
During inference, we only pass a high density surgical instrument image $\mathbf{I}$ as input to our model. The outputs are predicted surgical instrument handle bounding boxes. We further execute a post processing operator $\mathcal{P}$ to remove the redundant predicted bounding boxes that share the horizontal regions more than a predetermined threshold $\tau$. Detailed descriptions of post processing operator can be found in the supplement. 
\section{Experiments}

\subsection{Evaluation Metrics}
\vspace{-1mm}
We employ Mean Absolute Error (MAE) and Root Mean Squared Error (RMSE) as evaluation metrics. The detailed formulation of these two counting metrics and other localization metrics can be found in the supplements. 

\begin{figure*}[t]
\centering
\includegraphics[width=0.85\textwidth]{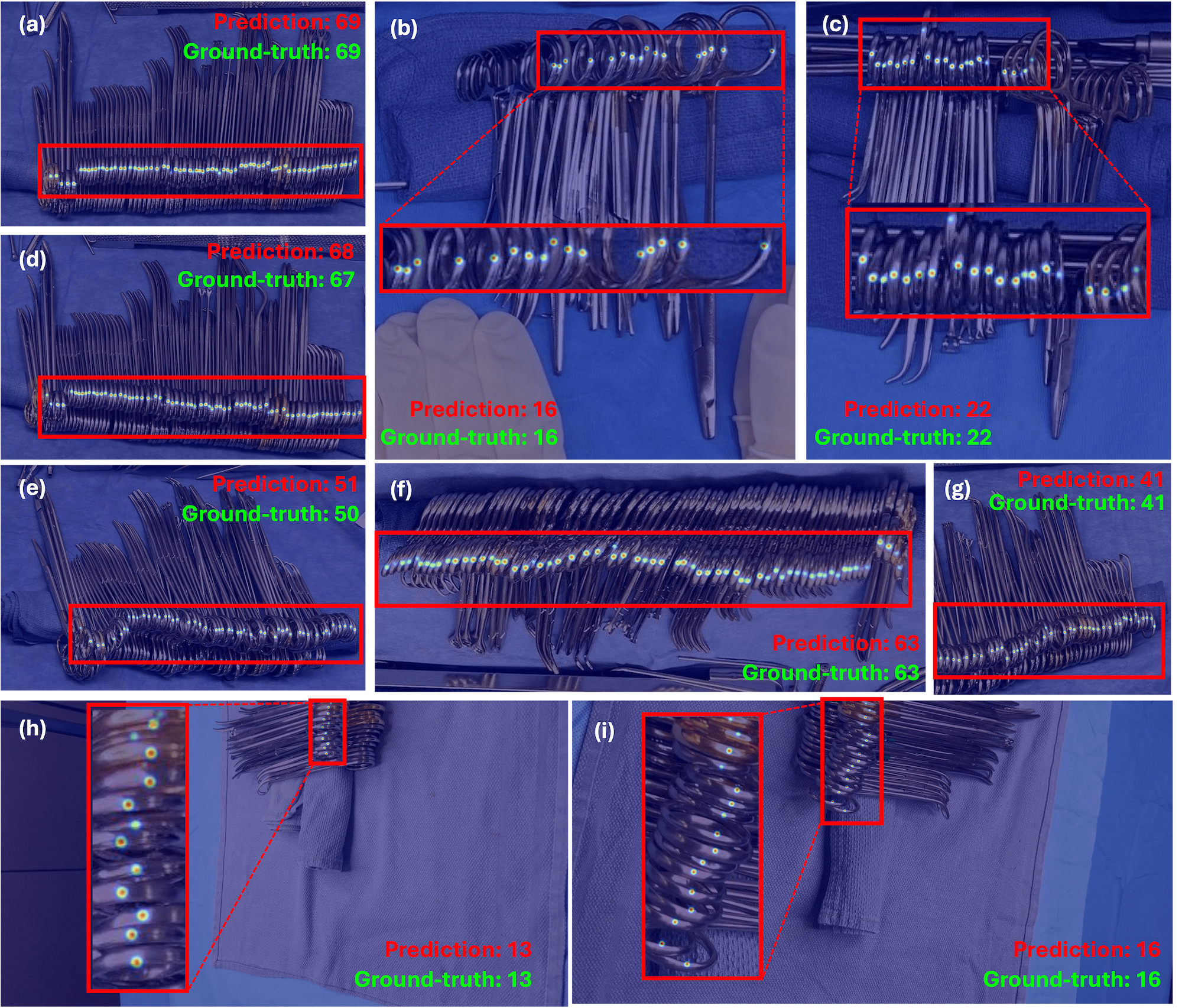} 
\vspace{-2mm}
\caption{\textbf{Qualitative results.} We present qualitative results from our CoLSR. Predicted surgical instruments number and ground-truth number are listed on each image. The detected surgical instrument handles are highlighted with laser points, which are also highlighted with red bounding boxes.}
\label{fig:qualitative results}
\end{figure*}

\begin{figure*}[t]
\centering
\includegraphics[width=0.87\textwidth]{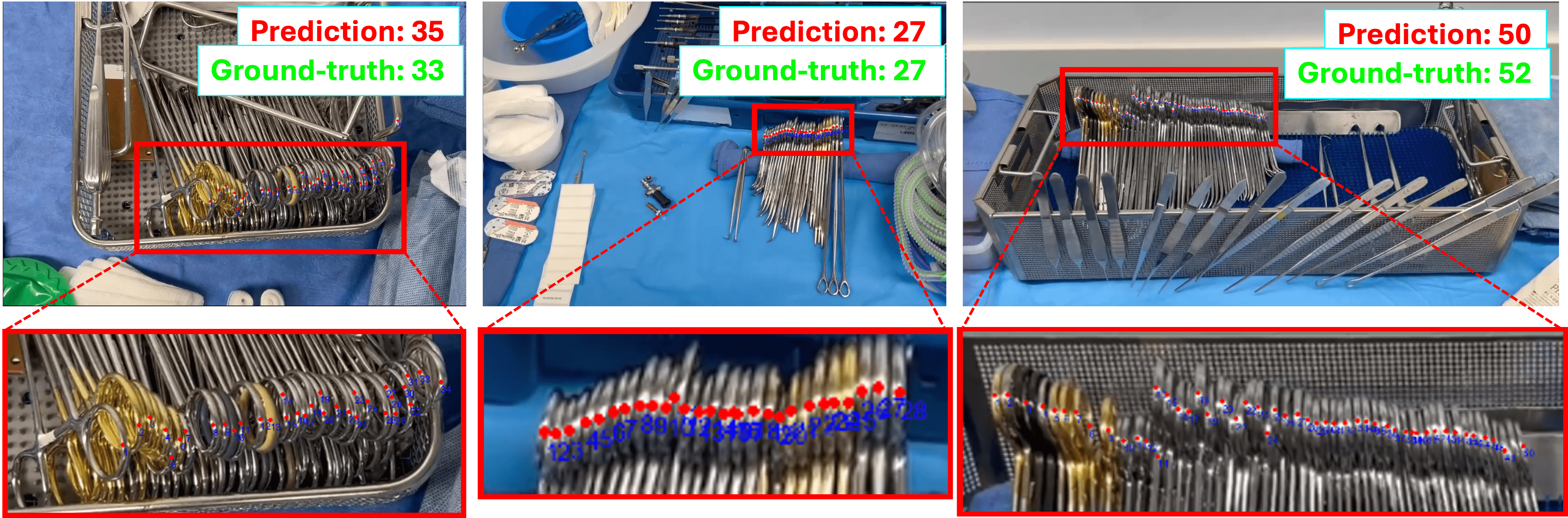} 
\vspace{-2mm}
\caption{\textbf{Generalization ability analysis.} We evaluate our model's generalization ability via in the wild images in operating rooms. The detected surgical instrument handles are highlighted with laser points, which are also highlighted with red bounding boxes.}
\label{fig:generalize_samples}
\end{figure*}


\vspace{-1mm}
\subsection{Quantitative Results}
\vspace{-1mm}
In Table~\ref{tab:compare_sota}, we compare the performance of CoLSR with state-of-the-art (SOTA) methods on the task of high-density surgical instrument counting. For a fair comparison, all SOTA counting baselines are finetuned on our SurgCount-HD dataset, except Qwen. CoLSR outperforms all competing methods in both MAE and RMSE metrics. The major reason lies in the primary limitation of existing counting methods, where they treat object instances as independent entities, lacking the spatial reasoning necessary to capture the dependencies and structural relationships among densely packed instruments. In contrast,  CoLSR explicitly models physical constraints, enabling it to reason over spatial arrangements and structural coherence more effectively. In addition, multimodality large vision-language models (MLVL) such as GPT5 \cite{gpt5} and Qwen-2.5-VL \cite{bai2025qwen2} also perform much worse compared with CoLSR, where the MAE of MLVLs are more than 10 times higher than CoLSR.

\begin{table}[t]
    \centering
    \resizebox{0.7\linewidth}{!}{
        \begin{tabular}{l c c} 
      \hline
      Method & MAE $\downarrow$ & RMSE $\downarrow$ \\ 
      \hline 
      CountGD \cite{amini2024countgd}  & 7.84 & 10.71 \\
      DQ-DETR \cite{huang2024dqdetrdetrdynamicquery} & 4.24 & 6.81 \\
      CrowdDiff \cite{ranasinghe2024crowddiff}  & 18.63 & 22.93 \\
      REC \cite{Dai_2024_CVPR} & 2.82 & 4.50 \\
      Qwen2.5-VL-7B-Instruct \cite{bai2025qwen2} & 17.06 & 21.72 \\
      \hline
      \hline
      \textbf{CoLSR} (Ours) & \textbf{0.88} & \textbf{1.27} \\
      \hline 
    \end{tabular}
    }
    \caption{Comparison with state-of-the-art methods, including: (1) counting and detection methods spanning detection-based (DQ-DETR), density-based (CountGD, REC), and diffusion-based (CrowdDiff) approaches; (2) multimodality large vision-language model (Qwen-2.5-VL).}
    \label{tab:compare_sota}
\end{table}

\begin{table}[t]
    \centering
    \resizebox{0.6\linewidth}{!}{
        \begin{tabular}{l c c} 
      \hline
      Method & MAE $\downarrow$ & RMSE $\downarrow$ \\ 
      \hline 
      $\Delta \mathcal{L}_{neigh}$ & 1.81 & 2.73 \\      
      $\Delta \text{CSL}$ & 2.05 & 3.30 \\
      $\Delta$Visual Exemplars & 1.5 & 2.21 \\
      $\Delta$Post & 0.996 & 1.48 \\
      \hline
      \textbf{CoLSR} (Full) & \textbf{0.88} & \textbf{1.27} \\
      \hline 
    \end{tabular}
    }
    \caption{\textbf{Ablation study results.} $\Delta \mathcal{L}_{neigh}$: without Neighboring Loss; $\Delta \text{CSL}$: without class-specific learnable prompts; $\Delta$ Visual Exemplars: without visual exemplars; $\Delta$Post: without post processing.}
    \label{tab:ablation}
\end{table}

\vspace{-1mm}
\subsection{Qualitative Results}
\vspace{-1mm}
Figure~\ref{fig:qualitative results} presents qualitative results of high-density surgical instrument counting using CoLSR. The visualizations highlight the robustness of our approach across various challenging scenarios, including  variations in camera angles (Figure~\ref{fig:qualitative results} (f), (h), (i)), instrument orientations (Figure~\ref{fig:qualitative results} (b), (c), (e)), and dense packing patterns (Figure~\ref{fig:qualitative results} (a), (d)).  

\begin{figure*}[t]
\centering
\includegraphics[width=0.93\textwidth]{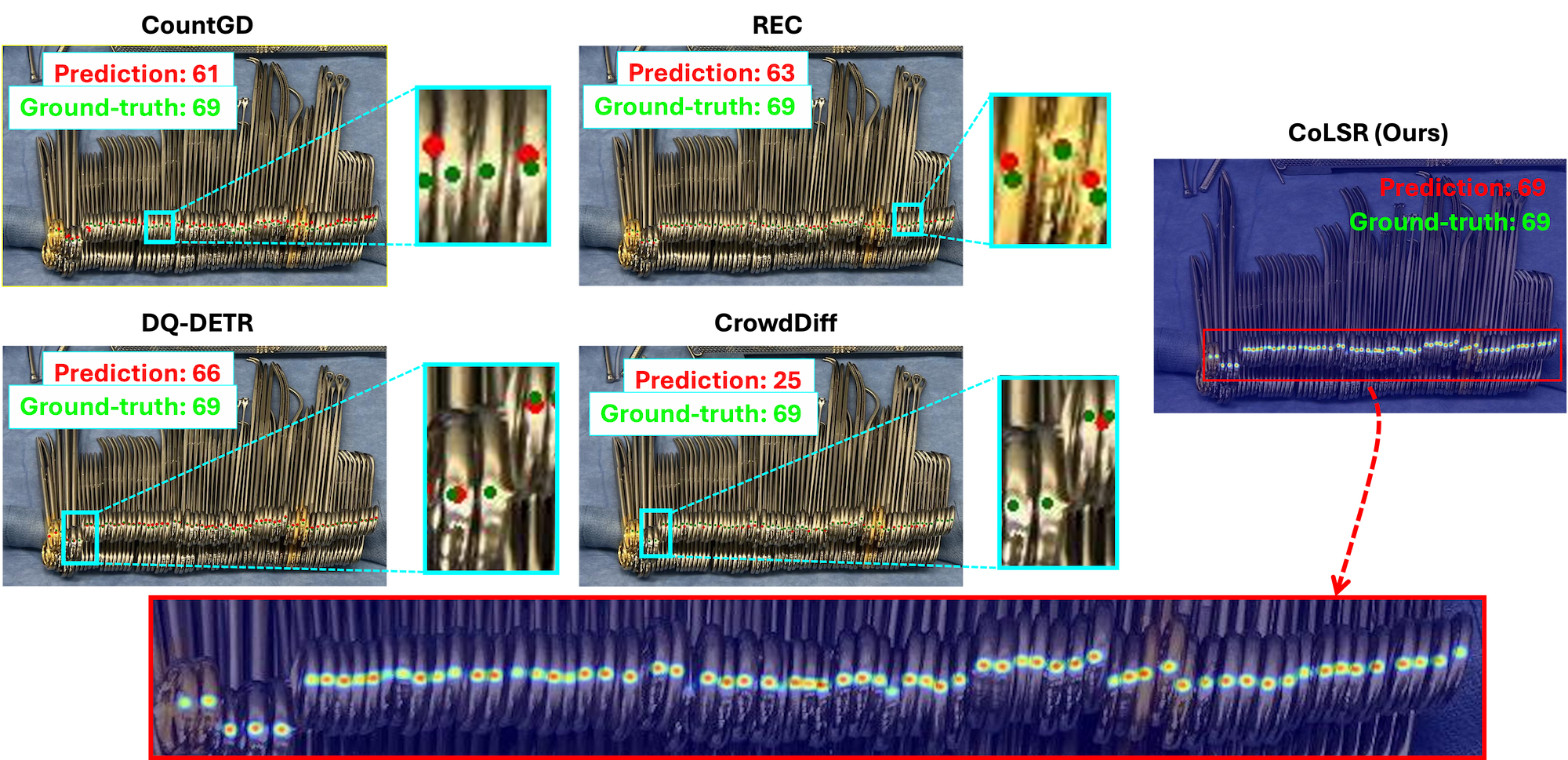} 
\vspace{-2mm}
\caption{\textbf{Comparison with SOTA methods.} Our CoLSR approach is compared with four existing SOTA methods for counting: CountGD, DQ-DETR, CrowdDiff and REC. For the four figures on the left side, \textcolor{green}{green} dots represent ground-truth, \textcolor{red}{red} dots represent predictions from different models.}
\label{fig:compare_sota}
\end{figure*}

Figure~\ref{fig:compare_sota} provides a visual comparison between our approach and existing SOTA methods for high-density instrument counting. As shown, SOTA methods often fail to detect all instrument handles, particularly in cluttered regions, resulting in under-counting. In contrast, CoLSR accurately localizes the instrument handles, as indicated by the cropped bounding boxes, demonstrating its effectiveness in handling densely packed scenes. To evaluate the generalization ability of our method, we test our model on in the wild images from operating rooms. Results in Figure ~\ref{fig:generalize_samples} indicate our method continually achieves robust results, demonstrating the generalization ability to real world operating room scenarios. We show more results in supplement material to evaluate the generalization ability of our method. 

\vspace{-1mm}

\subsection{Inference Speed}
\vspace{-1mm}
Our model is lightweight and achieves fast inference, running over \textit{\textbf{100$\times$ faster}} than manual human counting. Our mobile application achieves a \underline{peak} end-to-end (E2E) latency (including pre-processing, inference, and post-processing) of only 0.32s, compared to 44s required for manual counting. \underline{Average} latency is $0.28\pm0.02$s for our mobile application versus $25.12\pm11.63$s for human counting. A comparison figure and demo video are included in the supplementary material for clearer visualization.
 
\subsection{Ablation Study}
\vspace{-1mm}

\begin{figure}[t]
\centering
\includegraphics[width=0.48\textwidth]{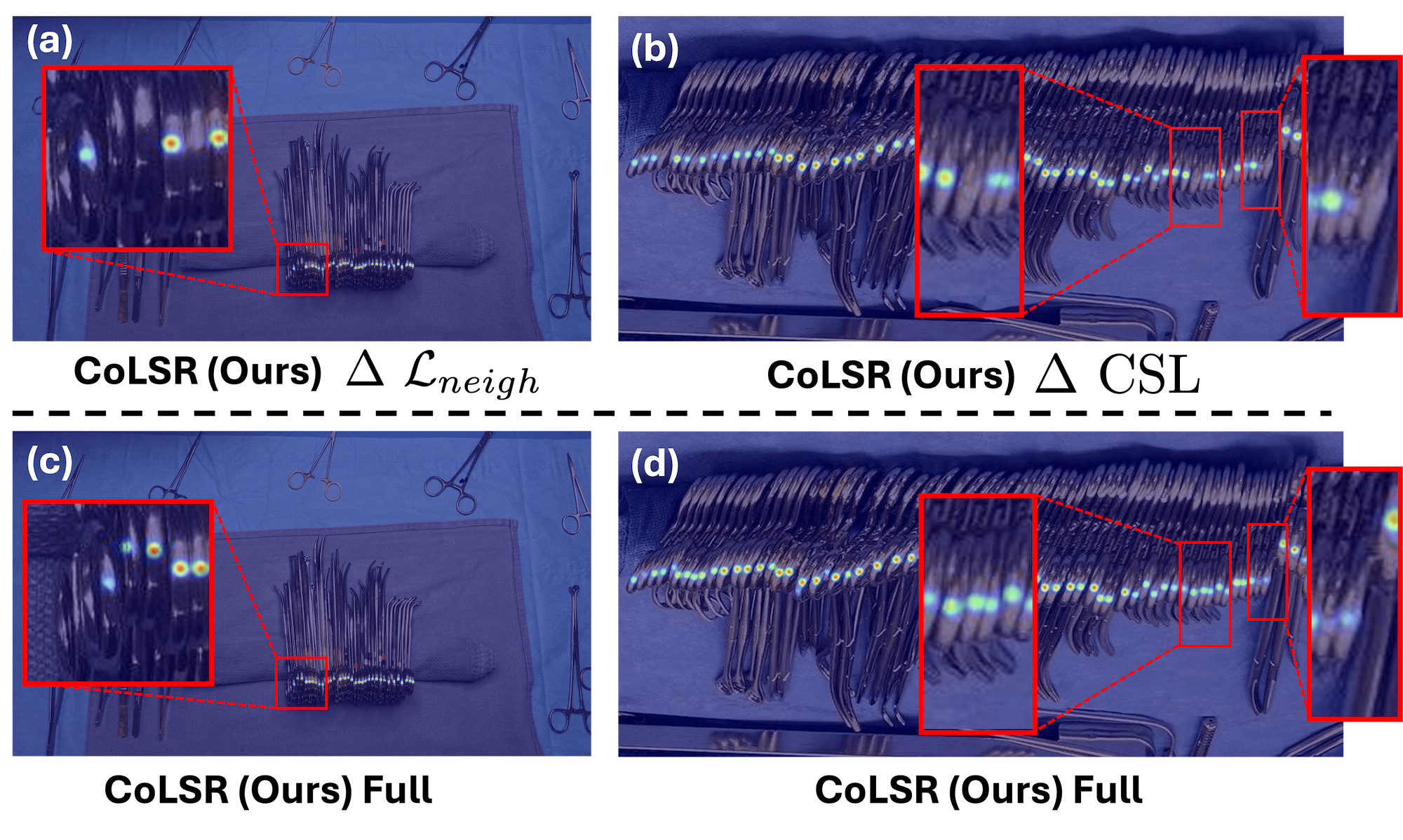} 
\vspace{-3mm}
\caption{\textbf{Ablation Studies.} \textbf{(a)} The highlighted region shows where the model failed to make correct predictions, indicating the model’s limited ability to form coherent visual chains. \textbf{(b)} Missed handles are mostly in areas with unclear boundary separation, making them harder to detect without class-specific learnable prompts. \textbf{(c, d)} Compared with the ablated results in (a) and (b), CoLSR effectively generates accurate predictions for the location of tightly packed surgical instrument handles.}
\vspace{-20pt}
\label{fig:ablation results}
\end{figure}

We conduct the following ablation studies to verify the effectiveness of each proposed component, including the neighboring loss, the class-specific learnable prompts and visual exemplars. 

\noindent \textbf{Effectiveness of neighboring loss function ($\Delta \mathcal{L}_{neigh}$ ).} Removing the neighboring loss diminishes the model’s ability to construct a visual chain for spatial reasoning, as shown in Fig.\ref{fig:ablation results} (a). This leads to a performance drop of approximately $105\%$ in terms of MAE, as shown in Table~\ref{tab:ablation}.

\noindent \textbf{Role of Learnable CSL Tokens ($\Delta$ CSL).} Eliminating the CSL prompts significantly impairs the model’s ability to detect fine-grained handle boundaries (further discussed in the supplementary material). This is further exacerbated when instruments are densely packed, causing the handle boundaries to appear merged as highlighted in Figure. \ref{fig:ablation results} (b). Consequently, the model’s performance degrades by approximately $133\%$ in terms of MAE, as shown in Table~\ref{tab:ablation}. \\ \\

\noindent \textbf{Pure Zero-shot training and inference ($\Delta$ Visual Exemplars).} As shown in Table~\ref{tab:ablation}, training and evaluation in a purely zero-shot setting without visual exemplars leads to a performance drop of approximately $70\%$ in terms of MAE.

\noindent \textbf{Role of postprocessing ($\Delta$ Post).} We remove postprocessing during inference and found slight drop of both MAE and RMSE in Table~\ref{tab:ablation}. 

\noindent \textbf{More ablation studies.} In supplementary, we show additional ablation studies including multi-loss weight selection, CSL token placement, among other variations.

\vspace{-1mm}
\subsection{Failure Analysis}
\vspace{-1mm}
In the supplementary material, we present representative failure cases. Most errors arise from the dense and visually ambiguous appearance of instruments in the images. Empirical results indicate that performance degrades in scenarios where gaps or occlusions disrupt the continuity of surgical instruments, making spatial reasoning more challenging. Potential solutions include leveraging multi-view inputs (e.g., short video sequences capturing multiple viewpoints) or incorporating depth information to better handle severe occlusions. As future work, we plan to integrate such diverse visual inputs into the counting pipeline to further enhance the robustness and reliability of our approach.

\vspace{-2mm}

\section{Conclusion}
\vspace{-2mm}
We introduce Chain-of-Look spatial reasoning framework that is inspired by human sequential counting behavior, designed to improve accuracy in densely packed surgical instrument scenes. By enforcing a structured visual chain and introducing a neighboring loss to model spatial constraints, our method outperforms existing SOTA counting models as well as multimodality large language models. This framework offers a generalizable approach that can be extended to broader applications requiring spatial reasoning in dense visual environments. Additionally, we introduce SurgCount-HD, a high-density surgical instrument dataset to facilitate benchmarking and drive future research in this domain.

\vspace{-2mm}

\section{Acknowledgment}
\vspace{-2mm}
The authors thank Brendan Fox, B.Sc., and Katy Tung, MD, for their contributions to data annotation. The authors also gratefully acknowledge Philip Seger, MD, for his work in data curation, annotation, and analysis. We further acknowledge Steven Schwaitzberg, MD, for his guidance in study design and resource allocation. Finally, we thank Gene Yang, MD, for his assistance with Institutional Review Board (IRB) processes.
All contributors are affiliated with the Jacobs School of Medicine and Biomedical Sciences.

{
    \small
    \bibliographystyle{ieeenat_fullname}
    \bibliography{main}
}

\clearpage
\setcounter{page}{1}
\maketitlesupplementary

\section{Analysis on Visual Chain Reasoning via Neighboring Loss}

\begin{figure*}[t]
\centering
\includegraphics[width=0.95\textwidth]{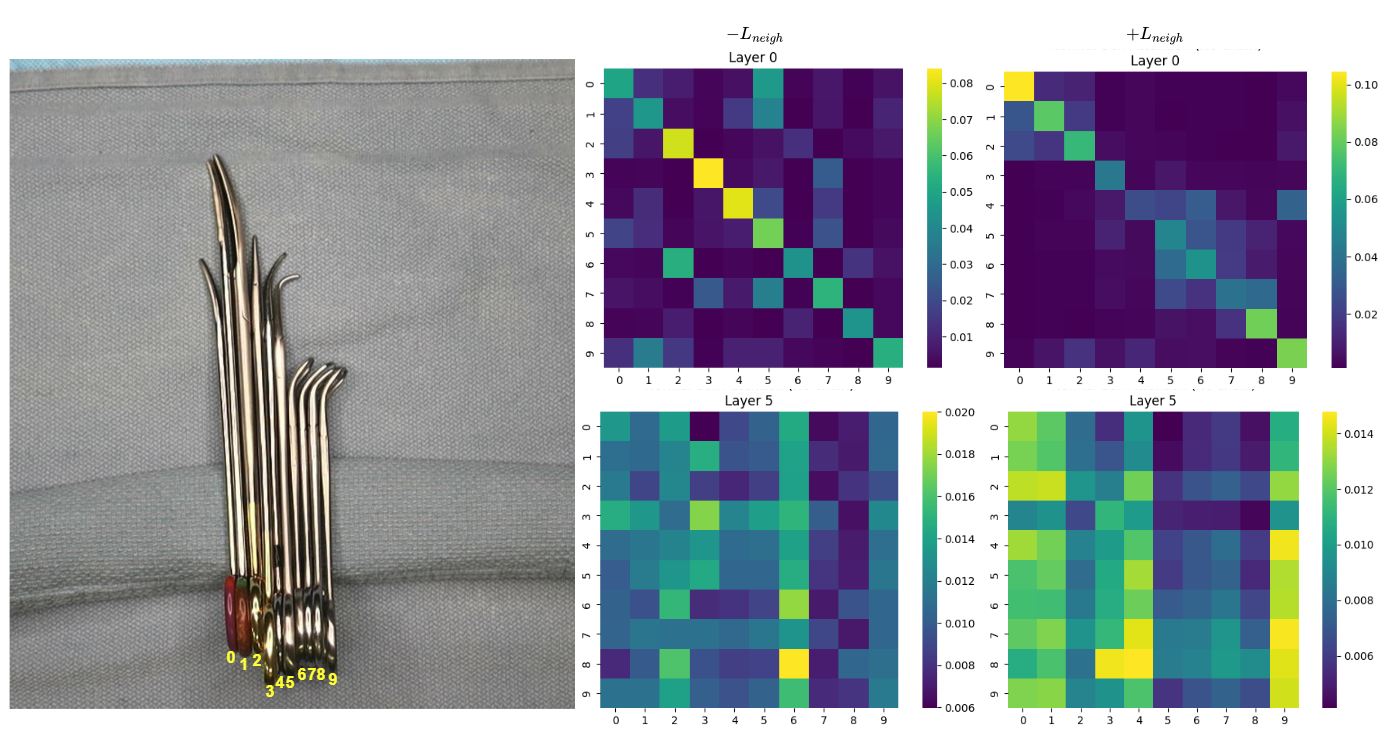} 
\caption{\textbf{Analysis on Chain-of-Look Visual Reasoning via Spatial Neighboring Loss.} Left: original surgical image. Right: attention maps from different decoder layers. ``$-L_{\text{neigh}}$'' denotes models trained without the Neighboring Loss, whereas ``$+L_{\text{neigh}}$'' indicates models trained with it. The visualizations show the self-attention outputs of the Cross-Modality Decoder, where each query corresponds to one surgical instrument (indexed 0–9). Queries and their associated attention distributions are ordered left-to-right according to the instrument labels in the original image. For each setting, we display attention maps from the first decoder layer (Layer 0), which primarily captures low-level spatial relationships, and from the final decoder layer (Layer 5), which reflects higher-level semantic focus.}
\label{fig:chainattn}
\end{figure*}

We demonstrate that the neighboring loss enforces a Chain-of-Look mechanism within the model’s reasoning process. Figure \ref{fig:chainattn} visualizes the self-attention scores of the query proposals in the Cross-Modality Decoder. At the first decoder layer (Layer 0), where the model primarily captures low-level spatial cues, we observe that removing the neighboring loss results in higher attention entropy, with focus spread across non-adjacent queries. In contrast, applying $\mathcal{L}_{\text{neigh}}$ constrains each query to attend mainly to its immediate predecessors and successors, forming a snake-like chained structure.

At the final decoder layer (Layer 5), the attention maps show that this chained behavior also shapes high-level semantic reasoning. For instruments that are densely clustered (labels 5–8), the model leverages the most visible and confident queries as anchors, reflected by the pronounced dark attention band, to improve the representation of uncertain and ambiguous queries. These structured interactions suppress hallucinations and ultimately improve counting accuracy.

\section{Implementation Details}
\vspace{-1mm}
We train the model for 30 epochs with a learning rate of \( 1 \times 10^{-4} \) using the Adam optimizer and a weight decay of \( 1 \times 10^{-4} \), which is reduced by a factor of ten after the $10^{th}$ epoch. Training is performed with a batch size of 4 on a single NVIDIA RTX 3090 GPU. The multi-loss weights are set as follows: \( \lambda_{\text{loc}} = 10 \), \( \lambda_{\text{neigh}} = 100 \), and \( \lambda_{\text{cls}} = 1 \). The number of CSL prompts used is 64, and the confidence threshold \( \sigma \) is set to 0.26. The rest of the training setup, including data pre-processing and augmentation strategies, follows the original CountGD \cite{amini2024countgd} configuration.

\section{CSL Prompt Design and Implementation}

\begin{figure*}[t]
\centering
\includegraphics[width=0.9\textwidth]{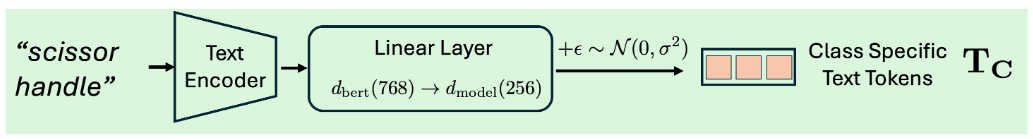} 
\caption{CSL Prompts Initialization with BERT Text Encoder}
\label{fig:cslinit}
\end{figure*}

The modified Feature Enhancer takes two different Class Specific Learnable (CSL) token instances, both initialized with the same text but diversified with Gaussian noise (Fig. \ref{fig:cslinit}). The first set of tokens are prepended to the concatenated set of visual exemplar and text tokens. We treat the CSL token as tunable text prompts, hence they are prepended to the latent tokens derived from text. These tokens, along with the image token, are used in the Bidirectional Attention module, where image-text and text-image cross-attention are computed. 

The second set of CSL token is prepended to the output of the bidirectional module and fed into the self-attention layer, where the self-attention between the text tokens is calculated. 

We introduce two sets of tokens to represent distinct functional roles: one captures the relation between text and image, while the other addresses text-specific nuances. The fused feature embedding $\mathbf{F}$ is constructed by concatenating these components:
\begin{equation}
    \mathbf{F}_{encoder} = [\mathbf{T}_{\text{CSL}} \,;\, \mathbf{T}_{\text{text}} \,;\, \mathbf{T}_{\text{vis}}] \in \mathbb{R}^{(l+2h) \times d},
\end{equation}
where $\mathbf{T}_{\text{CSL}} \in \mathbb{R}^{l \times d}$ denotes the $l$ CSL tokens, while $\mathbf{T}_{\text{text}}$ and $\mathbf{T}_{\text{vis}}$ (both $\in \mathbb{R}^{h \times d}$) represent the text and visual exemplar tokens, respectively. Following standard prompt-tuning methodology, we discard $\mathbf{T}_{\text{CSL}}$ before passing the sequence to the decoder.
\begin{equation}
    \mathbf{F}_{decoder} = [\mathbf{T}_{\text{text}} \,;\, \mathbf{T}_{\text{vis}}] \in \mathbb{R}^{(l+h) \times d}
\end{equation}

\subsection{Value Function for Hungarian Matching}
We utilize the CountGD formulation to define the Hungarian matching value function $v(i,k)$ between prediction $i$ and ground-truth $k$. Given $\alpha=0.25, \gamma=2$:

\begin{equation}
\small 
\begin{aligned}
    v(i,k) &= \underbrace{\|\mathbf{b}_{i} - \mathbf{b}_{k}\|_{1}}_{\textbf{bbox cost}} + \underbrace{\sum_{j=1}^{C} \tilde{y}_{kj} \left[\mathcal{L}_{\text{pos}}(p_{ij}) - \mathcal{L}_{\text{neg}}(p_{ij})\right]}_{\textbf{cls cost}}, \\
    \text{s.t.} \quad & \mathcal{L}_{\text{pos}}(p) = -\alpha(1-p)^{\gamma}\log(p+\varepsilon), \\
                      & \mathcal{L}_{\text{neg}}(p) = -(1-\alpha)p^{\gamma}\log(1-p+\varepsilon), 
\end{aligned}
\end{equation}

\noindent where $\mathbf{b}$ represents box center coordinates, $\tilde{y}_{kj}$ is the normalized target label for class $j$, and $p_{ij}$ is the predicted probability.

\section{Inference Speed}
\begin{figure}[t]
\centering
\includegraphics[width=0.48\textwidth]{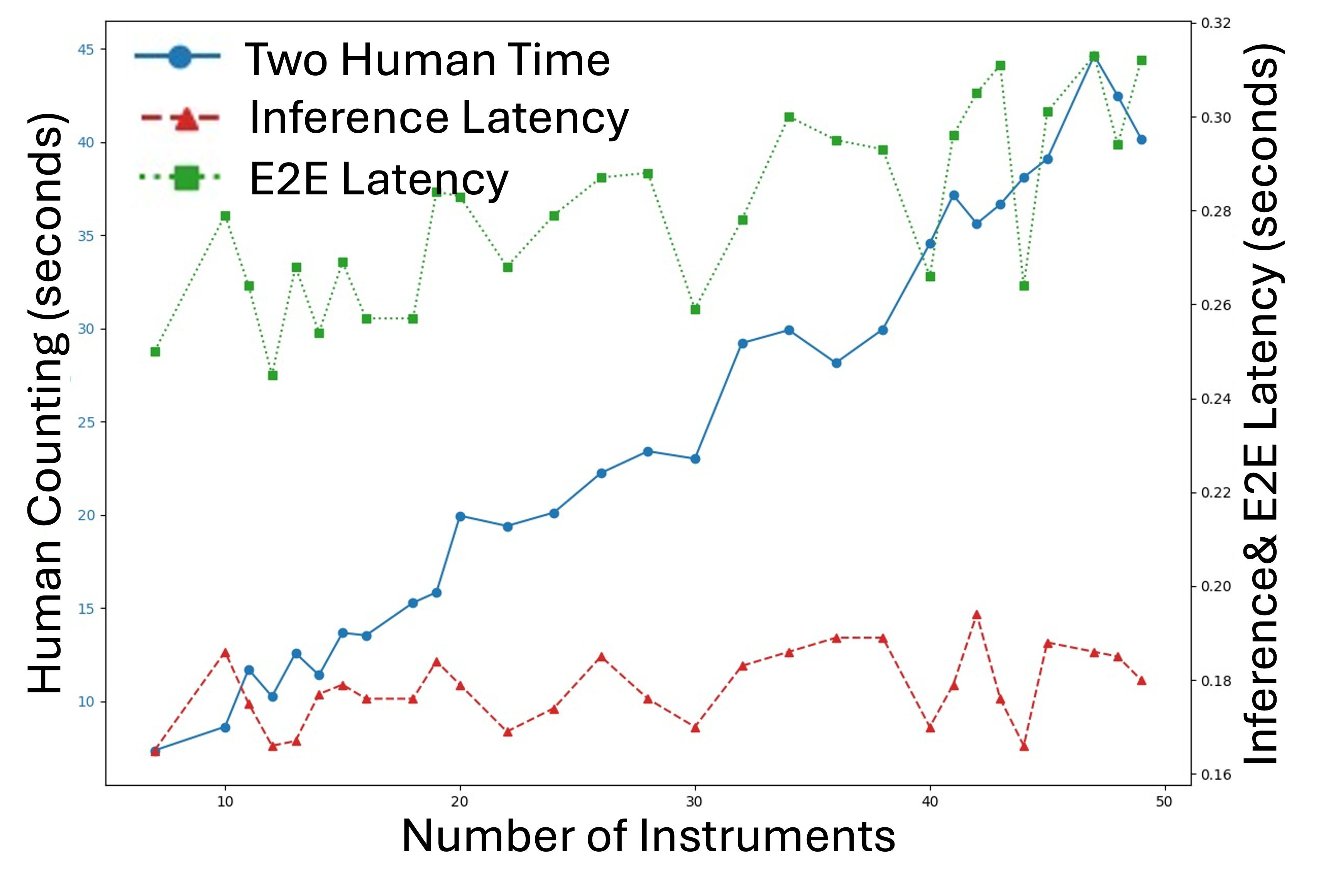} 
\caption{Time comparison between human counting and our model}
\label{fig:latency_time}
\end{figure}

Our model is lightweight and achieves fast inference, running over \textit{\textbf{100$\times$ faster}} than manual human counting. The experiments were performed across a range of scenarios, with the number of surgical instruments varying from 7 to 49 per trial. In each trial, two individuals performed manual counts, followed by a count using the mobile application. Figure~\ref{fig:latency_time} illustrates the contrast in performance between the traditional method and the app-based approach, highlighting the real-time efficiency gains enabled by the proposed system.

\section{Extended Ablation Results}

\subsection{CSL Prompts Placement: Appending vs. Prepending}
Previous studies \cite{li2021prefix} have highlighted how prompt placement affect transformer models. Our analysis (Table. \ref{tab:ablation1}) reveals that prompt placement significantly impacts performance, with prepending yielding 32\% better MAE than appending. Attention analysis shows prepended prompts maintain 3× stronger coupling with image features, while gradient flow analysis indicates on average 177× stronger supervision signals, explaining the performance disparity.

\begin{table}[t]
    \centering
    \resizebox{1\linewidth}{!}{
        \begin{tabular}{l c c c c} 
      \hline
      \textbf{Placement} & \textbf{MAE} $\downarrow$ & \textbf{RMSE} $\downarrow$ & \textbf{CSL Token Grad} $\uparrow$ & \textbf{Vision Attention Weights} $\uparrow$ \\
      \hline 
      \hline
      Append & 1.30 & 1.98 & Total: 0.044 & Mean: 0.00225 \\ 
      & & & Avg: 0.0037 & Std: 0.00325 \\
      Prepend & 0.88 & 1.27 & Total: 7.876 & Mean: 0.00643 \\
      & & & Avg: 0.656 & Std: 0.00498 \\
      \hline
    \end{tabular}
    }
    \caption{\textbf{Prompt Placement} Performance comparison across CSL prompt placements. See Section \ref{tab:promptmetrics} for metric definitions.}
    \label{tab:ablation1}
\end{table}

\subsubsection{Metrics Definitions}
\label{tab:promptmetrics}
\textbf{CSL Token Gradient Norms}: We measured the gradient norms for both types of CSL tokens-text and fusion-across all six encoder layers. For each type, the gradients were averaged over the layers using a single training batch to assess their relative contribution during backpropagation. \\
\textbf{Vision Multi-Head Attention Weights at Fusion Module}: To analyze the influence of prompts on visual attention, we extracted attention weights from the fusion module. Prompt-related weights were stacked and averaged across all four attention heads, followed by averaging over the batch dimension. This process was repeated for each of the six encoder layers, and the resulting layer-wise averages were further averaged to obtain a final mean value. The same training batch was used for both prompt placement configurations to ensure consistency.

We further investigate the impact of prompt initialization when prompts are prepended. Table \ref{tab:ablation4} shows that using task-specific initialization leads to notable performance improvements.

\begin{table}[t]
    \centering
    \resizebox{0.7\linewidth}{!}{
        \begin{tabular}{l c c c c c} 
      \hline
      \textbf{Initialization Type} & \textbf{MAE} & \textbf{RMSE} \\
      \hline 
      \hline
      Random & 1.32 & 1.96 \\
      Semantic (”scissor handle”) & 0.88 & 1.96 \\
      \hline
    \end{tabular}
    }
    \caption{\textbf{Prompt Initialization Strategy} Prepending task-specific initialized CSL prompts yields better performance compared to random initialization.}
    \label{tab:ablation4}
\end{table}

\subsection{LoRA versus CSL Tokens}

We explore whether adding explicit spatial conditioning using learnable prompt tokens offers benefits over weight adaptation methods in our instrument counting task. To test this, we inserted LoRA \cite{hu2022lora} adapters at the fusion and text encoder layers, mirroring the placement of CSL tokens. The adapter configuration ($\alpha$=32, Rank =16) was chosen to match the parameter count of the CSL tokens, allowing for a fair comparison.

\begin{figure*}[t]
\centering
\includegraphics[width=0.30\textwidth]{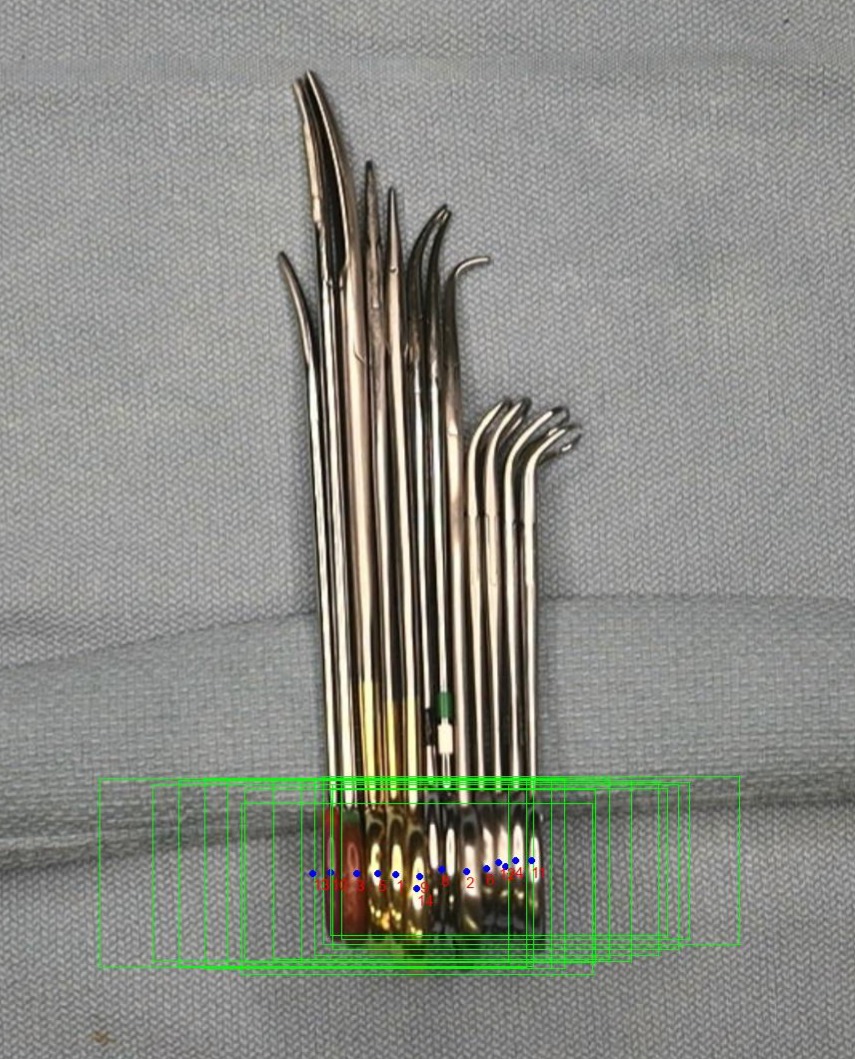}
\caption{Predicted bounding boxes using the LoRA method. Boxes are noticeably oversized and misaligned.}
\label{fig:lorasupp}
\end{figure*}

\subsubsection*{LoRA Parameters}
\begin{itemize}
\item Rank=16, $\alpha$=32
\item LoRA Params Per Layer : 16 (Rank) × 256 (In Features) + 16 (Rank) × 256 (Out Features) = 8,192 parameters
\item Text Encoder layers: 6 layers × 2 LoRA modules × 8,192 Per Layer $\approx$ 98K parameters
\item Fusion layers: 6 layers × 2 LoRA modules × 8,192 $\approx$ 98K parameters
\item Total LoRA Parameters $\approx$ 196K parameters
\end{itemize}

\subsubsection*{CSL Parameters}
\begin{itemize}
\item Text Encoder layers: 64 (CSL Token) × 256 (feature dim) × 6 layers $\approx$ 98K parameters
\item Fusion layers: 64 (CSL Token) × 256 (feature dim) × 6 layers $\approx$ 98K parameters
\item Total CSL Parameters $\approx$ 196K parameters
\end{itemize}

Our comparison between CSL Tokens and LoRA shows that token-level spatial conditioning leads to superior object detection performance despite LoRA's parameter efficiency. In our experiments, LoRA struggled with instrument localization, often missing center points and producing inaccurate bounding boxes (Figure. \ref{fig:lorasupp}). This suggests that LoRA's weight-space adaptation may lack the direct spatial conditioning beneficial for precise object localization in our setting. In contrast, the contrastive learning capability shown with CSL tokens (Section \ref{tab:section3}) appears to improve spatial reasoning that goes beyond parameter efficiency considerations. Our findings suggest that for this spatially sensitive detection task, explicit spatial conditioning through prompt tokens may provide capabilities that our constrained low-rank weight modification approach could not achieve.

\begin{table}[t]
    \centering
    \resizebox{1\linewidth}{!}{
        \begin{tabular}{l c c c c} 
      \hline
      \textbf{Method} & \textbf{MAE} $\downarrow$ & \textbf{RMSE} $\downarrow$ & \textbf{Mean L2 (matched)} $\downarrow$ & \textbf{Mean IoU (matched)} $\uparrow$ \\
      \hline 
      \hline
      LoRA & 5.63 & 7.66 & 10.42 & 0.028 \\
      CSL Tokens & 0.88 & 1.27 & 6.38 & 0.290 \\
      \hline
    \end{tabular}
    }
    \caption{\textbf{Counting \& Localization Metrics: LoRA vs. CSL Tokens.} The Mean IoU is the average IoU of all the matched bounding boxes in the test set.}
    \label{tab:ablation2}
\end{table}

\subsection{Multi-Loss Weight Selection}
Our method incorporates three distinct loss functions: Cross-Entropy Loss ($\mathcal{L}_{cls}$), Distance Loss ($\mathcal{L}_{loc}$), and Neighboring Loss ($\mathcal{L}_{neigh}$). We measured gradient norms across key shared model layers (encoder, decoder, fusion, text) for three loss weighting configurations to validate our $\lambda$ selection strategy.

\begin{table}[t]
    \centering
    \resizebox{1\linewidth}{!}{
        \begin{tabular}{l c c c c c} 
      \hline
      \textbf{($\lambda_{cls}, \lambda_{loc}, \lambda_{neigh}$)} & \textbf{$\mathcal{L}_{cls}$ Grad} & \textbf{$\mathcal{L}_{loc}$ Grad} & \textbf{$\mathcal{L}_{neigh}$ Grad} & \textbf{MAE} & \textbf{RMSE} \\
      \hline 
      \hline
      (1, 1, 1) & 0.915 & 0.0005 & 0.0002 & 3.23 & 4.20 \\
      (1, 10, 100) & 0.737 & 0.0007 & 0.0003 & 2.35 & 3.59 \\
      \textbf{(1, 10, 100)} & \textbf{28.50} & \textbf{0.055} & \textbf{0.106} & \textbf{0.88} & \textbf{1.27} \\
      \hline
    \end{tabular}
    }
    \caption{\textbf{Gradient Magnitude Analysis} Multi-Loss scaling factor selection.}
    \label{tab:ablation3}
\end{table}

As shown in Table. \ref{tab:ablation3} there exists a high imbalance in gradient magnitudes between the cross-entropy (CE) loss and auxiliary losses, with the latter exhibiting gradients 1,000–4,000× weaker under equal weighting. As a result, auxiliary objectives are effectively ignored during training. By introducing a loss weighting configuration of $\lambda$ = (1, 10, 100), we observe a substantial increase in auxiliary contribution (0.56\% vs. 0.09\%) while preserving CE dominance. This leads to a 30× increase in total gradient activity, enabling more expressive multi-objective optimization.

\section{CSL Prompts Effect and Contrastive Feature Learning}
\label{tab:section3}
\begin{figure*}[t]
\centering
\includegraphics[width=0.97\textwidth]{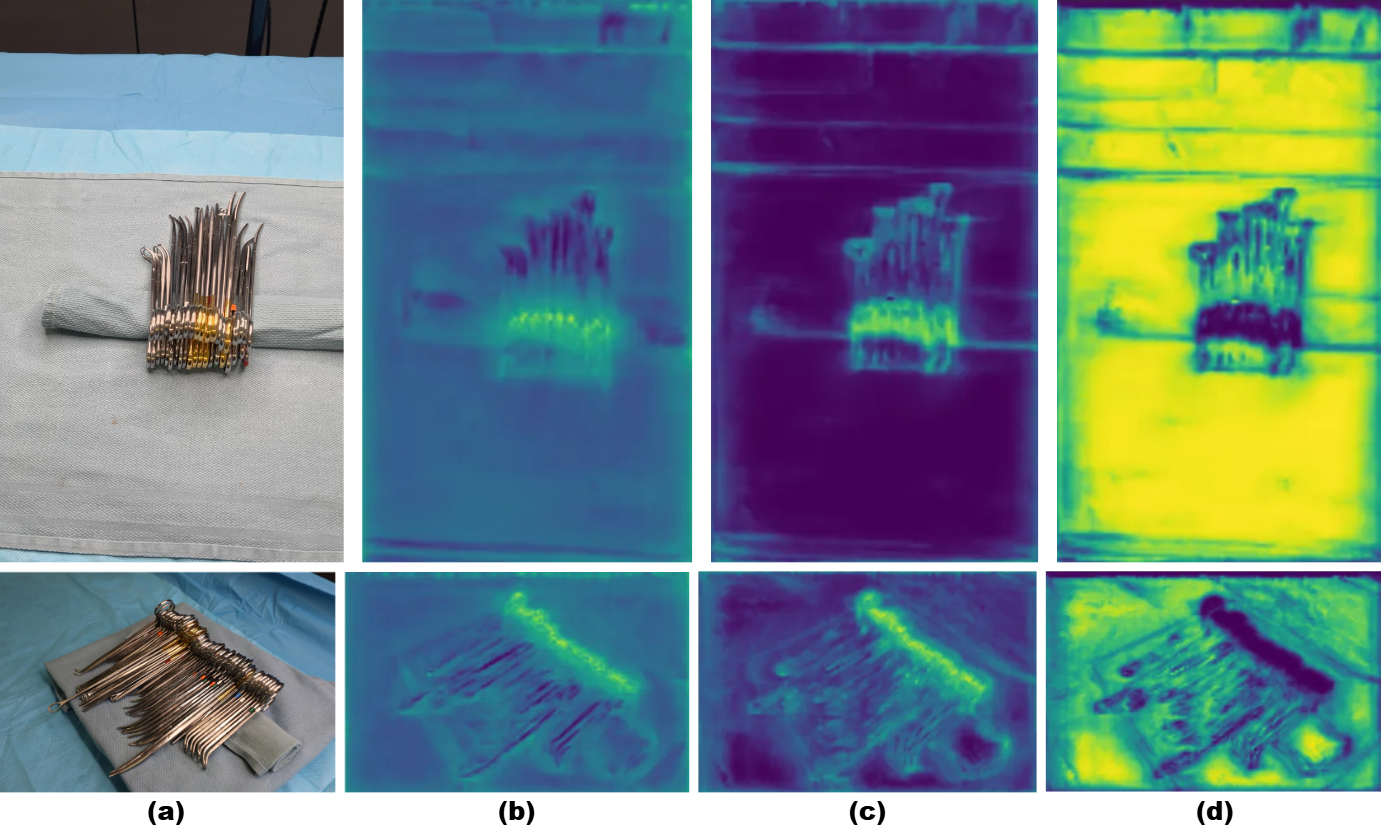}
\caption{\textbf{a)} Original Input Image \textbf{b)} Image-Text Attention Map extracted from the Feature Fusion Block - Without CSL Prompts \textbf{c)} Image-Text Attention Map when trained with CSL Prompts \textbf{d)} Image-CSL Token Attention Map}
\label{fig:cslsupp}
\end{figure*}

When trained without CSL prompts, the model’s attention is spread across and less focused on the handle regions, as illustrated in Fig. \ref{fig:cslsupp}(b–c). In contrast, CSL tokens learn contrastive features, as shown in Fig. \ref{fig:cslsupp}(d), where the attention on the handle is minimal. This complementary negation helps the text tokens to attend to the handle regions more precisely.
We experimented with varying numbers of CSL prompts \{16, 32, 64, 128\}, and found that 64 prompts produced the best performance based on the MAE metric.

\section{Determining Instrument Orientation for Neighboring Loss}
\label{subsec:orientation}
Firstly, we extract the center points from the predicted bounding boxes. Using these center points, we compute the difference between the maximum and minimum coordinates along the x- and y-axes. The axis with the largest difference is considered the dominant orientation of the instruments.
\begin{equation} \label{eq:my_equation_final_math_defs}
  \begin{split}
      1 - & \operatorname{int}\left(\left(P_{x}^{\text{max}} - P_{x}^{\text{min}}\right) > \left(P_{y}^{\text{max}} - P_{y}^{\text{min}}\right)\right)\\ &=
  \begin{cases}
    0 & \text{for} \ \  x\text{-axis} \\
    1 & \text{for} \ \  y\text{-axis} ,
  \end{cases}
  \end{split}
\end{equation}
where:
\begin{equation}
    \begin{split}
        &P_{\text{pred}} = \{(x_i, y_i) \mid \text{point } i \text{ is predicted}\}, \\
  &P_x = \{x_i \mid (x_i, y_i) \in P_{\text{predicted}}\} \quad \text{with } P_x \subseteq [0, W], \\
  &P_y = \{y_i \mid (x_i, y_i) \in P_{\text{predicted}}\} \quad \text{with } P_y \subseteq [0, H],
    \end{split}
\end{equation}
$W$ and $H$ denote the weight and height of the image. 

\section{Evaluation Metrics}

\subsection{Counting Metrics}
\subsubsection{MAE, RMSE}
We use the standard Mean Absolute Error (MAE) and the Root Mean Squared Error (RMSE) to measure as evaluation metrics. 

\begin{equation}
    \begin{split}
        \text{MAE} &= \frac{1}{N} \sum_{i=1}^{N} |N_P - N_G|, \\ \text{RMSE}&= \sqrt{\frac{1}{N} \sum_{i=1}^{N} (N_P - N_G)^2},
    \end{split}
\end{equation}
where N is the number of image samples, $N_P$ is the predicted count and $N_G$ is the ground truth count for image $N_i$.

\subsubsection{Grid Average Mean Absolute Error}
We also measure the Grid Average Mean Absolute Error (GAME) \cite{guerrero2015extremely} to evaluate the spatial accuracy of the predicted counts within each image. GAME quantifies how well the counting predictions are localized across subdivided regions of the image.
Given that surgical instruments in our dataset are typically concentrated within a limited spatial area, the GAME scores tend to decrease as the grid resolution parameter L increases (Table. \ref{tab:GAME_eval}). This is due to the presence of numerous grids containing no instruments, which contribute zero error to the overall score.
Of the 228 images in our test set, only 98 include instance-level annotations suitable for spatial evaluation. Therefore, the GAME scores and localization metrics were calculated exclusively on this subset.

Moreover, we also use detection-related counting metrics such as precision, recall and F1-score as defined in Equation. \ref{eq:metrics}.

\begin{table}[h!]
\centering
\resizebox{0.95\linewidth}{!}{
\begin{tabular}{lccc}
\hline
Method &  GAME-L1 $\downarrow$ & GAME-L2 $\downarrow$ & GAME-L3 $\downarrow$  \\
\hline
\hline
CountGD \cite{amini2024countgd} & 1.01 & 0.41 & 0.14 \\
REC \cite{Dai_2024_CVPR} & 0.60 & 0.25 & 0.08 \\
DQ-DETR \cite{huang2024dqdetrdetrdynamicquery} & 0.68 & 0.25 & 0.07 \\
\hline
\textbf{CoLSR (Ours)} & \textbf{0.54} & \textbf{0.23} & \textbf{0.07} \\
\hline
\end{tabular}}
\caption{GAME scores (L1, L2, L3) for different methods.}
\label{tab:GAME_eval}
\end{table}

\subsection{Localization Metrics}
Since the number of predicted instrument locations may not match the ground truth (GT) annotations, computing localization accuracy is non-trivial. To address this, we first filter predictions by selecting only those whose center points fall within any GT bounding box. These filtered predictions are then matched to GT points. In cases where multiple predictions fall within the same GT box, we apply the Hungarian algorithm using L2 distance as the cost function to perform one-to-one matching. \\
Unmatched predictions are treated as missed detections, while matched pairs are used to compute localization metrics. Specifically, for each image, we calculate the mean L2 distance (average localization error), the median L2 distance (typical error at the 50th percentile), and the 95th percentile of L2 distances (representing the worst 5\% of matched localizations). To obtain a single dataset-level metric, we take the mean of these three values across all images (Table. \ref{tab:localization_metrics}).
A similar procedure is applied for computing the Mean IoU reported in the Table. \ref{tab:ablation2}.

\begin{equation}
\label{eq:metrics}
    \begin{split}
    &\textbf{Steps for a single input}: \\
     &P_{\text{filtered}} = { p \in P_{\text{pred}} \mid \exists b \in B_{GT} \text{ such that } p \in b } \\
     &M^* = \underset{M}{\text{argmin}} \sum_{(p, g) \in M} |p - g|_2 \\
     &d_i = |p_i - g_i|_2 \\
     &\bar{d} = \frac{1}{N} \sum_{i=1}^{N} d_i \\
     &\text{Median Error} = \text{median}(d_1, d_2, \dots, d_N) \\
     &\text{95th Percentile Error} = P_{95}(d_1, d_2, \dots, d_N) \\ 
     &\text{True Positive (TP)} = N\\
     &\text{False Positive} = len(P_\text{pred}) - TP\\
     &\text{False Negative} = len(G_{GT}) - TP\\
     \end{split}
\end{equation}
\begin{equation*}
\begin{split}
&\text{where}: \\
        &P_{\text{pred}}:  \text{The set of all predicted center points.} \\
        &B_{GT}: \text{The set of all ground truth (GT) bounding boxes.}   \\
        &G_{GT}:  \text{The set of all GT center points.} \\
        &P_{\text{filtered}}:  \text{The set of all filtered center point prediction.} \\
        &M^*:  \text{The optimal one-to-one matching} \\
        &d_i:  \text{The L2 distance for the i-th matched pair} (p_i, g_i) \\
        &N: \text{The total number of matched pairs.} \\
    \end{split}
\end{equation*}

\section{Post Processing Operator}
\begin{figure}[t]
\centering
\includegraphics[width=0.2\textwidth]{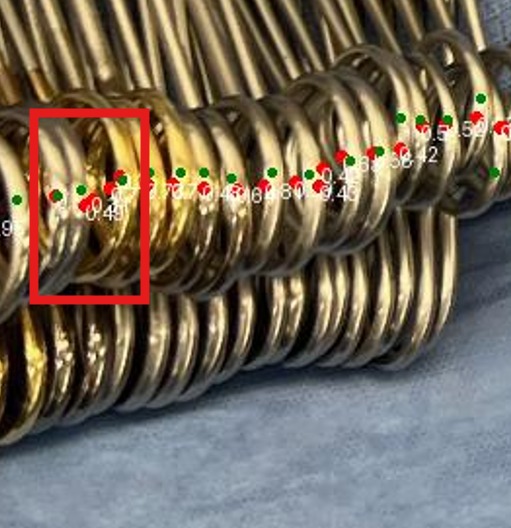} 
\caption{Example of duplicate points highlighted}
\label{fig:preproc}
\end{figure}

Due to the dense and ambiguous appearance of the instruments in the images, the model frequently produces multiple duplicate detections close to each other (Fig. \ref{fig:preproc}). To mitigate this, we applied a post-processing step to eliminate such points.

First, we sort the detected center points from left to right or top to bottom based on their orientation (Section \ref{subsec:orientation}). For each detected point, we examine neighboring points within a distance threshold $\theta$ along the given axis. If multiple points are found within this range, we retain only the point with the highest confidence score and discard the others, Alg: \ref{alg:preproc}.

\begin{algorithm}
\caption{Point Selection with Distance Threshold}
\label{alg:preproc}
\begin{algorithmic}[1]
\FOR{$P_i, P_j \in \{left, right\}$}
    \IF{$|P_i - P_j| < d$}
        \STATE $P_{selected} \gets \arg\max_{P \in \{P_i, P_j\}} \text{conf}(P)$
        \STATE $P_{removed} \gets \arg\min_{P \in \{P_i, P_j\}} \text{conf}(P)$
        \STATE Remove $P_{removed}$ from set
    \ENDIF
\ENDFOR
\end{algorithmic}
\end{algorithm}
\section{Divide and Conquer Inference}

\begin{figure*}[t]
\centering
\includegraphics[width=0.97\textwidth]{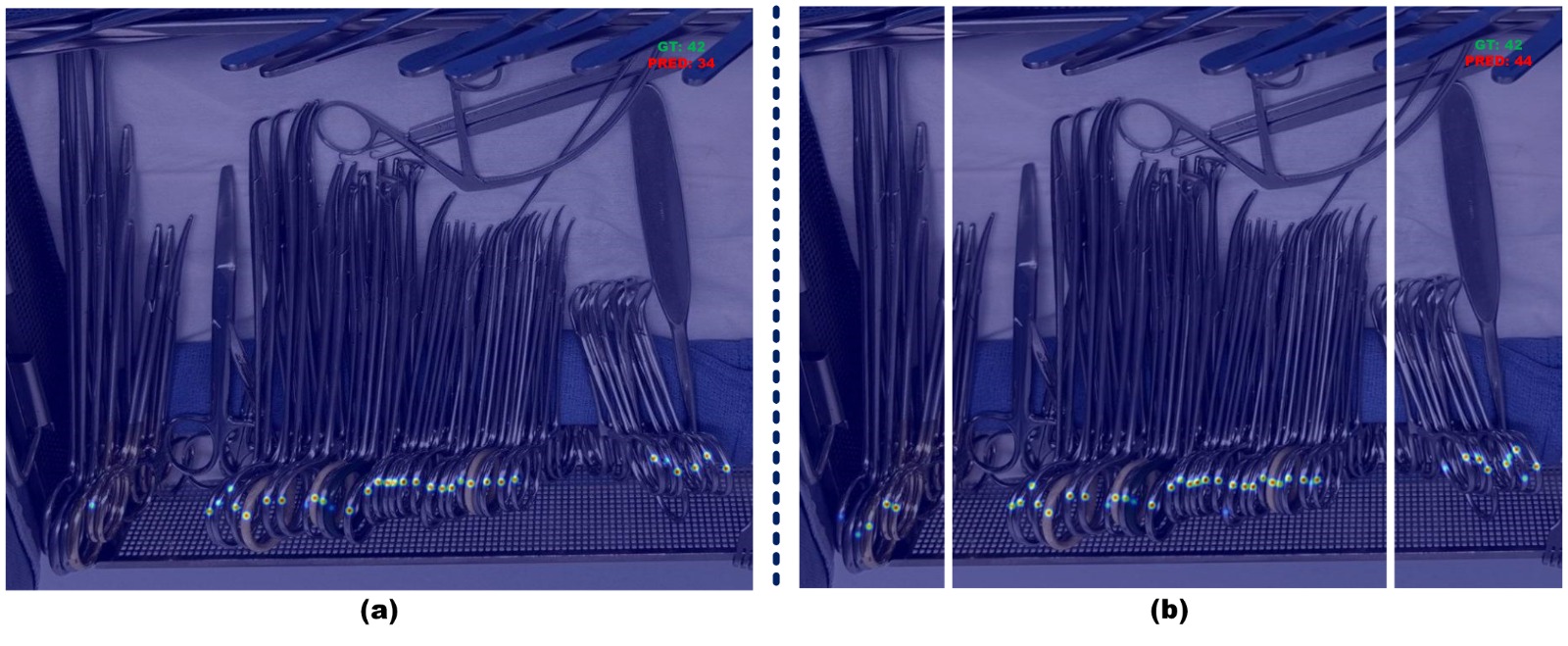} 
\caption{\textbf{a)} Prediction with single-pass inference \textbf{b)} Prediction with Divide-and-Conquer approach}
\label{fig:dandc}
\end{figure*}

CoLSR is designed to handle densely packed instrument clusters, which are the most common setup in real-world surgeries. However, its performance degrades when multiple dense clusters are spatially separated (Fig.\ref{fig:dandc}-a). This is due to the visual chain constraint enforced by the neighboring loss, which fails to capture long-range dependencies in such cases. To address this, we follow a two-stage approach, the Divide-and-Conquer strategy (Alg: 2,3).
In the first stage, the entire image is processed by the network, and the predicted center points are sorted along the x- or y-axis, depending on the instrument orientation (as described in \ref{subsec:orientation}).

We then compute the L2 norm between neighboring center points. If the distance between two neighbors exceeds $\delta$, the points on the left are grouped into one, and those on the right into another. This process is repeated until all center points are assigned to clusters.
Each cluster is then cropped from the original image and second-stage inference is performed independently on each dense region. Finally, the predictions of all the clusters are stitched together to produce the final output (Fig.\ref{fig:dandc}-b).

\begin{table}[h!]
\centering
\resizebox{1\linewidth}{!}{
\begin{tabular}{lcccc}
\hline
 & CountGD \cite{amini2024countgd} & REC \cite{Dai_2024_CVPR} & DQ-DETR \cite{huang2024dqdetrdetrdynamicquery} & \textbf{CoLSR (Ours)} \\
\hline
\hline
Mean L2 distance $\downarrow$ & 12.79 & 6.89 & \textbf{5.84} & 6.43 \\
Mean of Median L2 distance $\downarrow$  & 12.01 & 6.33 & \textbf{5.46} & 5.99 \\
Mean of 95th-Percentile L2 distance $\downarrow$  & 21.05 & 12.66 & \textbf{10.56} & 11.44 \\
\hline
Precision $\uparrow$ & 0.41 & 0.73 & 0.84 & \textbf{0.85} \\
Recall $\uparrow$ & 0.41 & 0.74 & 0.81 & \textbf{0.84} \\
F1 score $\uparrow$ & 0.41 & 0.74 & 0.83 & \textbf{0.85} \\
\hline
\end{tabular}}
\caption{Comparison of localization metrics results across different methods.}
\label{tab:localization_metrics}
\end{table}

\section{Limitations of Generalization}
\begin{figure*}[t]
\centering
\includegraphics[width=0.82\textwidth]{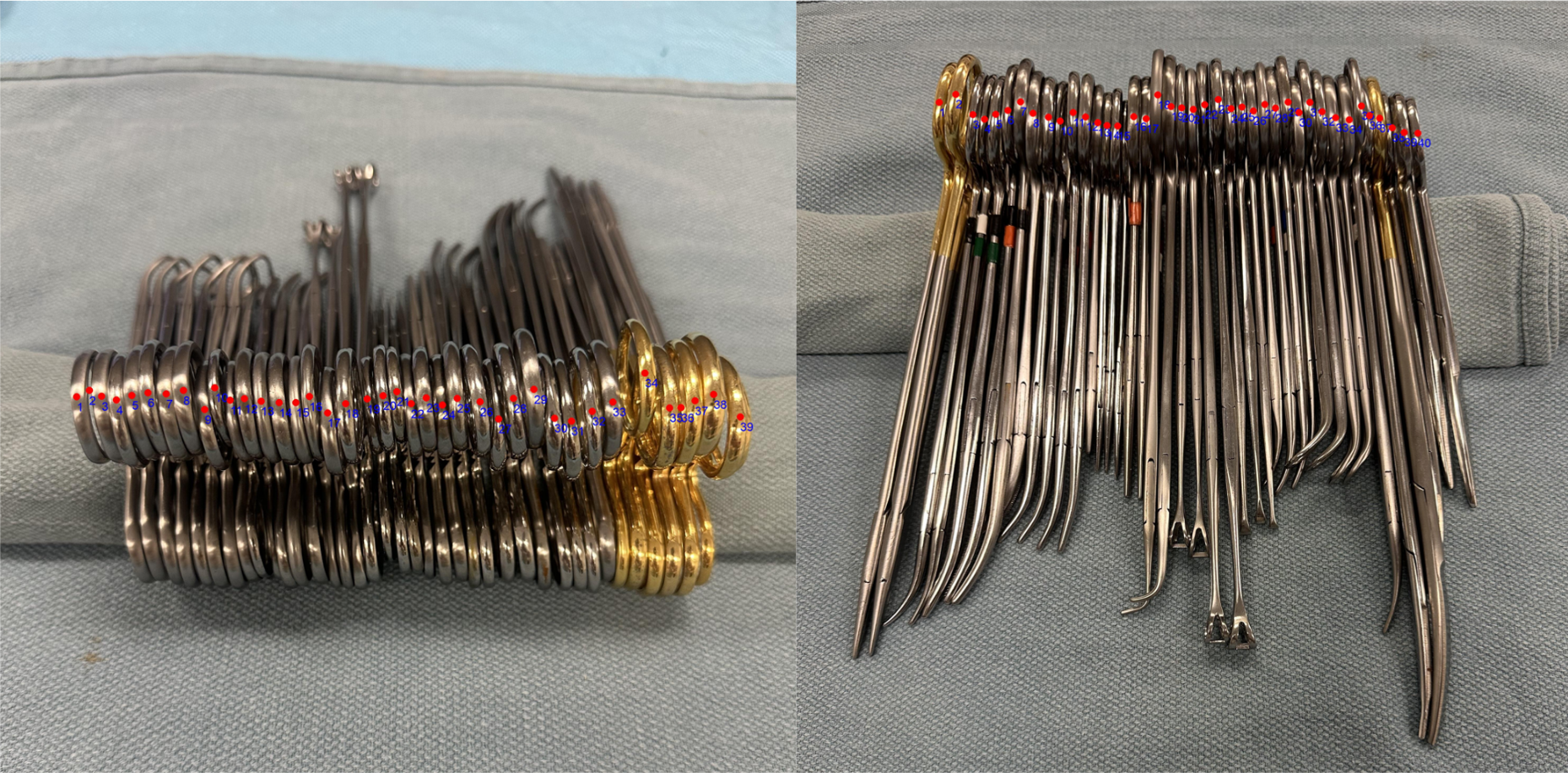}
\caption{\textbf{Robust inference samples captured from multiple angles.}}
\label{fig:general}
\end{figure*}

While our method demonstrates robustness to variations in angle and lighting conditions typical of operating room (OR) environments (Figure. \ref{fig:general}, Main Paper - Figure. 7), the scope of this work is limited to surgical instrument counting, as indicated by the paper title. Consequently, generalization to other domains may require further investigation.

\begin{table*}[t!]
\centering
\begin{tabular}{p{0.48\linewidth} p{0.48\linewidth}}

    \hrule height 1pt \vspace{2pt}
    \textbf{Algorithm 2:} Distance-Based Clustering 
    \hrule \vspace{2pt}
    \begin{algorithmic}[1]
    \STATE $clusters \gets \{\}$
    \STATE $cluster \gets [0]$ \COMMENT{cluster start}
    \FOR{$i = 0$ to $|pred\_points| - 2$}
        \STATE $p_i \gets pred\_points[i]$
        \STATE $p_{i+1} \gets pred\_points[i+1]$
        \IF{$\|p_i - p_{i+1}\|_2 > \delta$}
            \STATE $cluster$.append($i$) \COMMENT{cluster end}
            \STATE $clusters$.append($cluster$)
            \STATE $cluster \gets [i+1]$ \COMMENT{next cluster start}
        \ENDIF
    \ENDFOR
    \STATE $slices \gets \text{slice\_image}(clusters)$
    \RETURN $slices$
    \end{algorithmic}
    \hrule 
    & 
    
    \hrule height 1pt \vspace{2pt}
    \textbf{Algorithm 3:} Two-Pass Counting 
    \hrule \vspace{2pt}
    \begin{algorithmic}[1]
    \STATE $pred\_points \gets \text{run\_inference}(image)$ \COMMENT{first pass}
    \STATE $slices \gets \text{create\_cluster}(pred\_points, \delta)$
    \STATE $final\_detections \gets \{\}$
    \FOR{$slice \in slices$}
        \STATE $pred\_points \gets \text{run\_inference}(slice)$ \COMMENT{second pass}
        \STATE $final\_detections$.append($pred\_points$)
    \ENDFOR
    \RETURN $final\_detections$
    \end{algorithmic}
    \hrule
    
\end{tabular}
\end{table*}

\FloatBarrier
\balance

\end{document}